\documentclass[runningheads]{llncs}

\usepackage{graphicx,epsfig}

\begin{document}

\title{GUIDE: Unifying Evolutionary Engines through a
Graphical User Interface}
\author{
Pierre {\sc Collet} \inst1
\and
Marc {\sc Schoenauer} \inst2 }
\institute{
Laboratoire d'Informatique du Littoral,
Universit\'e du Littoral - C\^ote d'Opale\\
\and
Projet Fractales, INRIA Rocquencourt\\
{\tt Pierre.Collet@univ-littoral.fr,  Marc.Schoenauer@inria.fr}
}
\maketitle

\thispagestyle{empty}
\begin{abstract}
Many kinds of Evolutionary Algorithms (EAs) have been described
in the literature since the last 30 years. However, though most of them
share a common structure, no existing software package allows the user
to actually shift from one model to another by simply changing a few
parameters, e.g. in a single window of a Graphical User Interface. This
paper presents GUIDE, a {\em \underline{G}raphical \underline{U}ser
\underline{I}nterface for \underline{D}REAM \underline{E}xpe\-riments}
that, among other user-friendly features, unifies all kinds of EAs into a
single panel, as far as evolution parameters are concerned. Such a window
can be used either to ask for one of the well known ready-to-use
algorithms, or to very easily explore new combinations that have not yet
been studied. Another advantage of grouping all necessary elements to
describe virtually all kinds of EAs is that it creates a fantastic
pedagogic tool to teach EAs to students and newcomers to the field.

\end{abstract}

\section{Introduction}
As Evolutionary Algorithms (EAs) become recognised as practical
optimisation tools, more and more people want to use them
for their own problems, and start reading some literature.
Unfortunately, it is today very difficult to get a clear picture of the
field from papers or even from the few books that exist.
Indeed, there is not even a common terminology between different
authors.

Many papers (see e.g.  \cite{HEC}), refer to the different
``dialects'' of Evolutionary Computation, namely Genetic Algorithms
(GAs), Evolution Strategies (ESs), Evolutionary Programming (EP) or
Genetic Programming (GP).

>From a historical perspective, this is unambiguous: broadly
speaking (see \cite{Fogel:Fossile:98} for more details),
 GAs were first described by J. Holland \cite{Holland} and popularised by
D.E. Goldberg \cite{Goldberg89} in Michigan (USA);
 ESs were invented by I. Rechenberg \cite{Rechenberg} and H.-P. Schwefel
\cite{Schwefel} in Berlin (Germany);
 L. Fogel \cite{Fogel-pere} proposed Evolutionary Programming on the
US West Coast;
 J. Koza started the recent root of Genetic Programming
\cite{Koza}.

However, when the novice reader tries to figure out the
differences between those dialects from a scientific perspective, she/he
rapidly becomes lost: from a distance, it seems that GAs manipulate
bitstrings, ESs deal with real numbers and GP handles programs represented
by parse-trees. Very good ---so the difference seems to lie in
{\em representation}, i.e.  the kind of search space that those dialects
search on.

But what about EP then ? Original EP talks about Finite State Automata,
but many EP papers deal with parametric optimisation (searching a subspace
of {\bf R}$^n$ for some $n \in$ {\bf N}) and many other search spaces;
and the ``real-coded GAs'' also do parametric optimisation, while inside
the GA community, the issue of representation is intensively discussed
\cite{Radcliffe:ICGA91,Whitley:Eurogen97,Surry:PhD98}, and many {\em
  ad hoc} representations are proposed for
instance for combinatorial problems
\cite{Radcliffe-variance}. Some ESs also deal with
combinatorial representations
\cite{Herdy:GECCO2002} or even
bitstrings \cite{Baeck-hyperbole}; and even in the field of GP, which
seems more
clearly defined by the use of parse-trees, some linear representations are
currently used \cite{Nordin:ICGA95}.

So our patient and persevering newcomer starts delving into the technical
details of the algorithms, and thinks he has finally found the fundamental
differences: those dialects differ by {\em the way they implement artificial
Darwinism}.
\begin{itemize}
\item Indeed, GAs use proportional or tournament-based selection, generate
as many offspring as there are parents, and the generation loop ends by
replacing all parents by all offspring.
\item In ESs, each of the $\mu$ parents generates a given number of offspring,
and the best of the $\lambda$ offspring (resp. the $\lambda$ offspring +
the $\mu$ parents) are deterministically selected to become the parents of
the next generation. The algorithm is then called a $(\mu,\lambda)-ES$
(resp. $(\mu+\lambda)-ES$).
\item EP looks very much like a $(P+P)-ES$, except that competition for
survival between parents and offspring is stochastic.
\item In GP, only a few children are created at each generation from some
parents selected using tournament-based selection, and they
replace some of the parents that are chosen using again some
tournament. But wait, this way of doing is precisely called \ldots
Steady-State GA ---so this is a GA, then !
\end{itemize}

And indeed, GP started as a special case of GA manipulating parse-trees
and not bitstrings (or real numbers or \ldots), and became a field of its
own because of some other technical specificities, but also, and mostly,
because of the potential applications of algorithms that were able to
create programs and not ``simply'' to optimise.

``Ah, so the differences come from the applications then ?'' will ask the
newcomer. And again, this will only be part of reality, as there
is in fact no definite answer: all points of view have their answer,
historical, technical, ``applicational'' or even \ldots political (however,
this latter point of view will not be discussed in this paper, devoted
to scientific issues).

But what does the newcomer want to know ? She/He wants to be able to use
the best possible algorithm to solve her/his problem. So she/he is certainly
not concerned with historical differences (apart from curiosity), and she/he
wants to find out first what exactly is an Evolutionary Algorithm, and
what different parameters she/he can twiddle to make it fit her/his needs.

Starting from the target application, a newcomer should first be able
to choose
any representation that seems adapted to the problem, being informed
of what ``adapted to the problem'' means, in the framework of
Evolutionary Computation: the representation should somehow capture
the features that seem important for the problem at hand. As this is
not the central issue of this paper, we refer the user to the
literature, from the important seminal work of Radcliffe
\cite{Radcliffe:ICGA91} to more recent trends in the choice of a
representation \cite{Bentley:GECCO99}.

The only other thing for which the user cannot be replaced is of
course the choice (and coding) of the fitness function to
optimise. But everything
else that a user needs to do to run an Evolutionary Algorithm
should be tuning some parameters, e.g. from  some graphical interface:
many variation operators can be automatically designed
\cite{Surry:PPSN96,Surry:PhD98}, and most implementations of artificial
Darwinism can be described in a general framework that only requires
fine tuning through a set of parameters.

This paper addresses the latter issue, with the {\em specification of any
evolution engine, unified within a single window} of a Graphical User
Interface named {\em GUIDE} (where {\em evolution engine} means the way
artificial Darwinism is implemented in an EA
in the selection and replacement steps).
In particular, this paper will {\bf not} mention any
representation-specific feature (e.g. crossover or
mutation), nor any application-specific fitness function.

Section \ref{EA} briefly recalls the basic principles
of  EAs, as well as the terminology used in this
paper. Section \ref{GUIDE}
presents the history of GUIDE, based on the specification language {\em
EASEA} \cite{Collet:PPSN2000}. Section \ref{engine} details the Evolution
Engine Panel of GUIDE, demonstrating that it not only fulfills the
unification of all historical ``dialects,'' but that it also
allows the user to go far beyond those few engines and to try
many more yet untested possibilities. Finally, section \ref{conclusion}
discusses the limitations of this approach, and sketches some future
issues that still needs to be addressed to allow a wide dissemination
of Evolutionary Algorithms.

\section{Basic principles and terminology}
\label{EA}
Due to the historical reasons already mentioned in the introduction,
even the terminology of EAs is not yet completely unanimously agreed
upon.
Nevertheless, it seems sensible to recall here both the basic
skeleton of an EA and the terminology that goes with it.
This presentation will however be very brief, as it is assumed
that the reader is familiar with at least some existing EAs, and will
be able to recognise what she/he knows.
Important terms will be written in boldface in the rest of the paper.

\subsection{The skeleton}
The goal is to optimise a
 {\bf fitness function} defined on a given search space, mimicking
the Darwinian principle that {\em the fittest individuals survive and
  reproduce}. A generic EA can be described by the following steps:

\begin{itemize}
\item {\bf Initialisation} of {\em population} $\Pi_0$, usually
  through a uniform random choice over the search space;
\item {\bf Evaluation} of the individuals in  $\Pi_0$
  (i.e. computation of their fitnesses);
\item {\bf Generation} $i$ builds population $\Pi_{i}$ from population
   $\Pi_{i-1}$:
\begin{itemize}
\item  {\bf Selection} of some parents from $\Pi_{i-1}$, biased by the
  fitness ({\em the ``fittest'' individuals reproduce});
\item Application (with a given probability) of
  {\bf variation operators } to the selected parents, giving birth to
  new individuals, the {\bf offspring};
Unary operators are called {\bf mutations} while n-ary operators are
called {\bf recombinations} (usually, $n=2$, and the term {\bf
  crossover} is often used);
\item {\bf Evaluation}  of newborn offspring;
\item {\bf Replacement} of population $\Pi_{i}$ by a new
  population that is created from the old parents of population
  $\Pi_{i-1}$ and the newborn offspring, through another round of
  Darwinian selection ({\em the fittest individuals survive}).
\end{itemize}

\item Evolution stops when some predefined level of performance has
  been reached, or after a given number of generations
  without significant improvement of the best fitness.
\end{itemize}


An important remark at this point is that such a generic EA
is made of two parts that are completely orthogonal:
\begin{itemize}
\item the {\em problem-dependent} components, including
the choice of the search space, or space of {\bf genomes}, together
with their  initialisation, the variation operators and of course the
fitness function.
\item on the other hand, the {\bf evolution engine} implements the
artificial Darwinism part of the algorithm, namely the selection and
replacement steps in the skeleton given above, and should be able to handle
populations of objects that have a fitness, regardless of the actual
genomes.
\end{itemize}

\subsection{Discussion}
\label{discussion}
This is of course a simplified view, from which many actual algorithms
depart. For instance, there are usually two search spaces involved in
an EA: the space of genomes, or {\bf genotypic space}, where the
variation operators are
applied, is the space where the actual search takes place; and the
{\bf phenotypic space}, where the fitness function is actually
evaluated. The mapping from the genotypic space onto the phenotypic
space is called the decoding of solutions, with the implicit
assumption that the solution the user is looking for is the phenotypic
expression of the genome given by the algorithm. Though the nature
of this mapping is of utter importance for the success of the
algorithm, it will not be discussed at all here, as GUIDE/EASEA only
considers genotypes, hiding the phenotypic space in the fitness.

Other variants of EAs do violate the pure Darwinian dogma, and hence
do not enter the above framework: many
variation operators are not ``blind,'' i.e. do bias their actions
according to fitness-dependent features;
conversely, some selection
mechanisms do take into account some phenotypic traits, like some
sharing mechanisms involving phenotypic distances between individuals
\cite{Sareni:TEC98}.

Nevertheless, we claim that the above generic EA covers a vast
majority of existing algorithms\footnote{It even potentially covers
  Multi-Objective EAs, as these ``only'' involve specific
  selection / replacement steps. However, MOEAs are not yet available
  in GUIDE, but will be soon.}, and most importantly, is a mandatory
step for anyone intending to use Artificial Evolution to solve a given
problem.
In that context, the existence of a user-friendly interface allowing
anyone with
little programming skills to design an Evolutionary Algorithm
following this skeleton is a clear dissemination factor for EC as an
optimisation technique: Such were the motivations for  EASEA
\cite{Collet:PPSN2000} and, later, for GUIDE.

\section{GUIDE overview}
\label{GUIDE}

GUIDE is designed to work on top of the EASEA language
({\em \underline{EA}sy \underline{S}pecification of
\underline{E}volutionary \underline{A}lgorithms})
\cite{Collet:PPSN2000}.
The EASEA language and compiler have been designed to simplify the
programming of
Evolutionary Algorithms by providing the user with all EA-related routines,
so that she/he could concentrate on writing the code that is
specific to the application, as listed in section \ref{EA}, i.e.:
the genome structure and the corresponding
initialisation, recombination, mutation and evaluation functions.

On the evolution engine side, a set of parameters allows the user to pick up
existing selection/replacement mechanisms, and was designed
to supersede most existing combinations, while allowing new ones.
One of the most important features of EASEA is that it is
library-independent: from an EASEA source code, the compiler generates
code for a target library, that can be either the C++ libraries
{\em GALib} \cite{GALib} and
EO \cite{EO:EA01},  or, more recently, the Java library JEO
\cite{JEO:GECCO02}.
The resulting code is then compiled using the routines from the
corresponding library ---but the user never has to dive into the
intermediate complex object-oriented code.

Unfortunately, whereas the goal of relieving the user from the tedious
task of understanding an existing EC library was undoubtedly
reached since the very first versions of EASEA, back in 1999,
the specification of the evolution parameters
implicitly supposes some deep knowledge of existing
evolution engines. Moreover, the lack of agreed terminology makes it
even difficult for EC-advanced users to pick up their favorite
algorithm: GA practitioners, for instance, have hardly heard of the
replacement step, because in ``standard'' GAs, the number of generated
offspring is equal to the number of parents, and all offspring simply
replace all parents (generational replacement).
The need for a graphical interface was hence felt necessary quite
early in EASEA history.


Such a graphical interface was eventually developed as part of the DREAM
European project  IST-1999-12679
\cite{DREAM:PPSN2002}
(Distributed Resource Evolutionary Algorithm Machine ---hence the name GUIDE),
whose evolutionary library JEO
\cite{JEO:GECCO02} is one of the
possible targets for EASEA compilation.

\subsection*{GUIDE: a quick tour}
In a programming environment, the Graphical User Interface is the entry
point at
the highest possible level of interaction and abstraction. In the GUIDE/EASEA
programming environment, the idea is that even a non-expert programmer should
be able to program an EA using one of the underlying libraries without
even knowing about it ! GUIDE must therefore at least allow the user to:
\begin{enumerate}
\item specify the numerous parameters of any evolutionary engine by way
of a point-and-click interface,
\item write or view the user code related to problem-specific operators,
\item compile the experiment by a simple click,
\item run the experiment, and visualise the resulting outputs in some
window(s).
\end{enumerate}

While the second to fourth points above are merely a matter of
implementation, and
by no way could justify a scientific paper in a conference,
the first point not only is original, but also clearly (graphically)
highlights the common features of most existing EC paradigms: the user
can specify {\em any} evolution engine within  a single window.

%

\begin{figure}\centering
\mbox{\epsfxsize=12cm\epsfbox{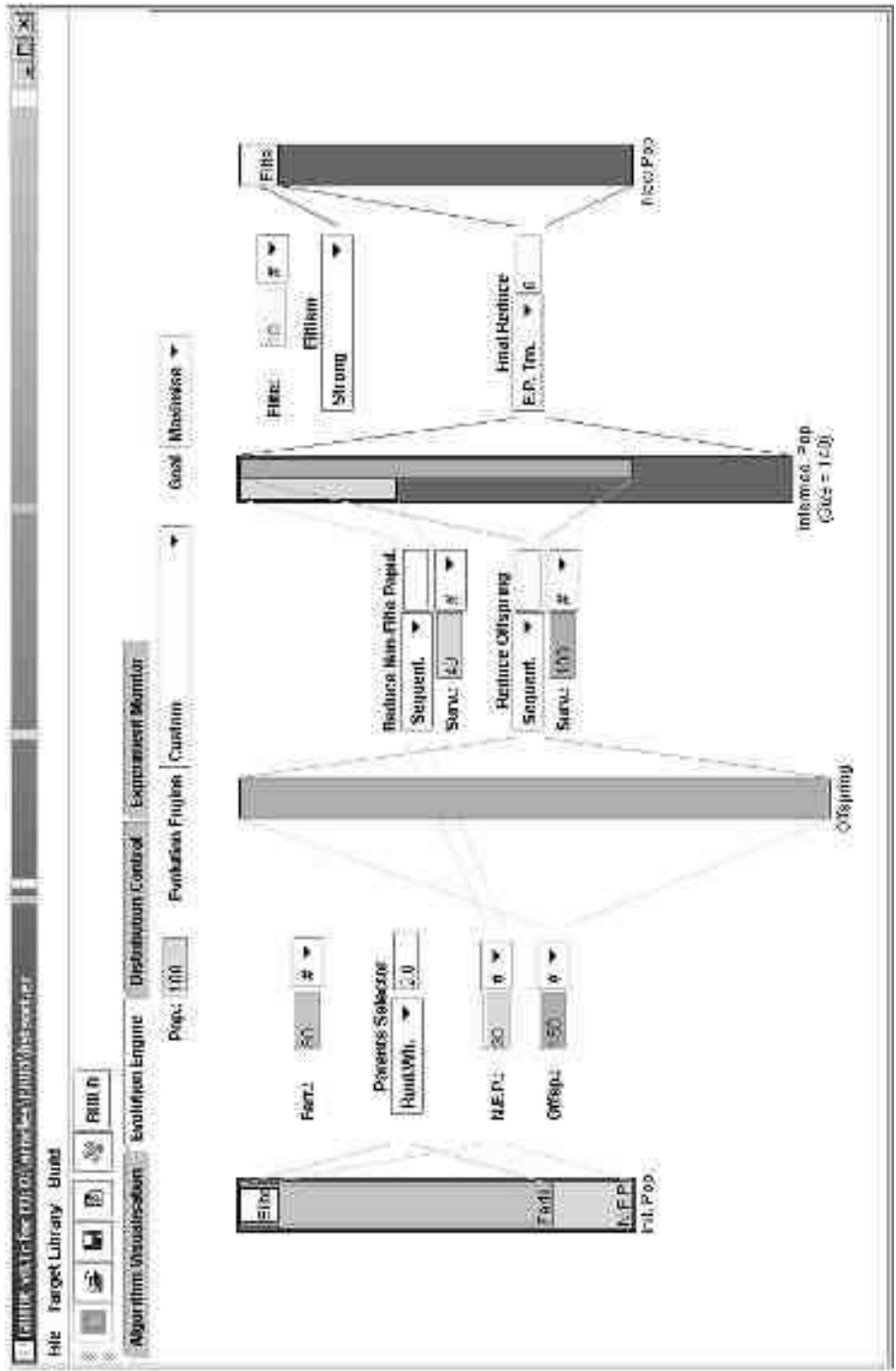}}
\caption{GUIDE: Evolution Engine Panel}
\label{evoleng}
\end{figure}

The structure of GUIDE reflects this point of view, and offers four
panels to the user (see the tags on Figure \ref{evoleng}):
\begin{description}
\item [Algorithm Visualisation Panel] to visualise and/or
modify pro\-blem-de\-pen\-dent code. It contains a series of text windows,
each  window referring to the equivalent ``sections'' of EASEA source
code \cite{Collet:PPSN2000} that  the user has to type in. This is where the
user writes the code for genetic operators such as initialisation, mutation,
recombination, and most importantly for the evaluation function.
\item [Evolution Engine Panel] for the Darwinian components. This panel will
be extensively described in section \ref{engine}.
\item [Distribution Control Panel] to define the way different islands
communicate in a distributed model \cite{DREAM:PPSN2002}. This panel will soon
be adapted to ParaDisEO \cite{Paradiseo:2003}, the recent \underline{Para}llel
\underline{Dis}tributed version of EO.
\item [Experiment Monitor Panel] to compile and run the
  experiment. At the moment, only compilation is possible from there.
The user has to run the program from outside GUIDE.
\end{description}

GUIDE obviously also offers (see Figure \ref{evoleng}) a top-menu bar from
where
the user can save/load
previous sessions to {\tt Files}, choose the {\tt Target Library} and
{\tt Build} the executable file ---and, as usual, some of these
actions can be fired by some icons from the toolbar.

\section{Evolution Engine Panel}
\label{engine}

This section describes the most innovative feature of GUIDE: the panel
where the user can specify the evolution engine,
either from pre-defined ``paradigmatic'' engines, or by designing
new combinations of selections and replacements fitting her/his taste.

\subsection{Evolution Engines}
The concept of evolution engine designates the two steps of the basic
EA that implement some artificial Darwinism:  {\bf selection} and
{\bf replacement}. Basically, both steps involve the ``selection'' of
some individuals among a population. However, at least one important
difference
between those steps makes it necessary to distinguish among them,
even when only one is actually used (like in traditional
GAs or ESs):  an individual can be selected {\em several times} for
{\em reproduction}
during the selection step, while it can be selected {\em at most once}
during the {\em replacement} step (in this case, the non-selected individuals
simply disappear).

Another important feature is implemented in a generic way in
GUIDE: {\bf elitism}. It will be discussed separately (section
\ref{elitism}).

\subsection{Panel description}
Figure \ref{evoleng} shows the Evolution Engine Panel of GUIDE. On top of the
panel, the user can specify the population size (here 100), whether
the fitness is to be maximised or minimised, and which type of
Evolution Engine should be used: this type can be any of the
existing known engines, described in next section \ref{paradigms}, or set
to {\tt Custom}, as in Figure  \ref{evoleng}.

Below are some vertical bars, representing the number of individuals
involved in different steps of a single generation: the left part
of these bars describes
the selection step (including, implicitly, the action of the
variation operators, but not their description), and starts with the
initial parent population ---the leftmost bar, labeled {\tt
  Init.$\:$Pop}.   The right
two bars specify  the replacement step, and end with the {\tt New
  Pop.}$\:$that will be the {\tt Init.$\:$Pop.}$\:$of next generation ---and
hence has the same height (number of individuals contained) that the
{\tt Init.$\:$Pop.}$\:$bar.


\subsubsection{Parameters input}
There are two kinds of user-defined parameters that completely specify
the evolution engine in GUIDE: the sizes of some intermediate
populations, and the type of some selectors used inside the selection
and the replacement steps. When a parameter is a type from a predefined
list, a pull-out menu presents the possible choices to the user.

All sizes in GUIDE can be set either graphically, using the mouse to
increase or decrease the size of the corresponding population, or
using the numeric pad and entering the number directly. In the latter
case, the number can  be entered either as an absolute value, or as a
percentage of the up-stream population size ---the choice is
determined by the small menu box with either the ``\#'' or the ``\%''
character.

\subsubsection{Selection}

The selection step picks up individuals from the parent population
{\tt Init.$\:$Pop.}$\:$and handles them to the variation operators to
generate the {\tt Offspring} population.
In GUIDE, the parameters for the selection step are:

\begin{itemize}
\item the type of
selector that will be used to pick up the parents,
together with its
parameters (if any).
Available selectors (see e.g. \cite{Chakraborty:JEC96}
for the definitions)are {\tt Roulette wheel} and (linear) {\tt
  ranking} (and an associated
selection pressure), {\tt deterministic
  tournament} (and the associated tournament size),
{\tt stochastic tour\-nament} (and the associated probability),
plus the trivial {\tt random} (uniform selection)
and {\tt sequential} (deterministic selection selecting from best to
worse individuals in turn).
\item The number of {\tt Fertile} individuals: non-fertile individuals
  do not enter the selection process at all. This is somehow
  equivalent to truncation selection, though it can be performed here
  before another selector is applied among the fertile individuals
  only.
\item The size of the {\tt Offspring} population.
\end{itemize}


The last field in the ``selection'' area of the panel that has not
been discussed here is the {\tt N.E.P.}, or \underline{N}on
\underline{E}lite \underline{P}opulation, whose size is that of the
population minus that of the {\tt Elite} and that will be discussed in
section \ref{elitism} below. At the moment, consider that this
population is the whole parent population.

In the example of Figure \ref{evoleng}, only the 80 best individuals
will undergo roulette wheel selection with selection pressure 2.0, and
150 offspring will be generated. (Although roulette wheel is known to have
several weaknesses, notably compared to the Stochastic Universal Sampling
described by Baker\cite{baker} it is still available in GUIDE,
 mainly because all underlying libraries implement it.)

\subsubsection{Replacement}

The goal of the replacement procedure is to choose which individuals
from the parent population {\tt Init.$\:$Pop.}$\:$and the offspring
population {\tt Offspring}  will build the  {\tt New
  Pop.}$\:$population.
In GUIDE, the replacement step is made of three {\bf
  reduction} sub-steps: a reduction is simply the elimination of some
individuals from one population according to some Darwinian {\bf reduce}
procedure:

\begin{itemize}
\item First, the {\tt Non Elite Population} (the whole
{\tt Init.$\:$Pop.}$\:$in the absence of elitism) is reduced (the user
must enter the
  type of reducer to be used, and the size of the reduced population).
\item Second, the {\tt Offspring} population is reduced (and again,
  the user must set the type of reducer and the size of the reduced
  population).
\item Finally, both reduced populations above are merged into
  {\tt Intermed.$\:$Pop.}$\:$(for intermediate population). The corresponding
  bar is vertically divided into two bars of different colors: the
  survivors of both populations. This
  population is
  in turn reduced into the final {\tt New Pop.}, whose size has to be
  the size of the parent
  population: hence, only the type of reducer has to be set there.
\end{itemize}

The available types of reducers include again the {\tt sequential} and
{\tt random} reductions, as well as the deterministic and stochastic
{\tt tournaments} (together with their respective parameters),
that repeatedly eliminate bad individuals. An additional option is
the {\tt EP tournament} (together with the tournament size $T$): in this
stochastic reducer,
each individual fights against $T$ uniformly chosen opponents, and
scores 1 every time it is better; the best total scores then survive.\\

Note that many possible settings of those parameters would result in
some unfeasible replacements
(e.g. asking for a size of {\tt Intermed.$\:$Pop.}$\:$smaller that the {\tt
  Pop.} size). Such unfeasible settings are filtered out by
GUIDE \ldots as much as possible.

In the example of Figure \ref{evoleng}, the best 40 individuals from
the {\tt N.E.P.}$\;$(the initial population in the absence of elitism)
are merged with the best 100 offspring,
and the resulting intermediate population is
reduced using an EP tournament of size 6.

\subsubsection{Elitism}
\label{elitism}
All features related to elitism have been left
aside up to now, and will be addressed here.
There are two ways to handle elitism in EC literature, termed
``strong'' and ``weak elitism'' in GUIDE. The
user first chooses either one from the menu in the replacement
section, and then  sets the number of {\tt Elite} parents in the
corresponding input box (setting the {\tt Elite} size to $0$ turns off
all elitism). \\

\noindent
{\bf Strong elitism} amounts to put some (copy) of the best initial
  parents in the {\tt New Pop.}$\:${\em before} the replacement step,
  without any selection against
  offspring whatsoever. The remaining ``seats'' in the {\tt New Pop.}$\:$are
  then filled with the specified replacement.
Note that the elite population nevertheless enters the
  selection together with the other parents ---and that, of course,
  only the {\tt N.E.P.}$\:$competes in the reduction leading to the parent part
  of the {\tt Intermed.$\:$Pop}.\smallskip

\noindent
{\bf Weak elitism}, on the other hand, only takes place {\em
    after} normal replacement, in the case when
  the best individual in the {\tt New Pop.}$\:$is worse than the best
  parent of the {\tt Init Pop}. In that case, all parents from the
  {\tt Elite} population that are better than the best individual of the
  {\tt New Pop.}$\:$replace the worst individuals in that final
  population. This type of elitism is generally used with an {\tt
    Elite} size of 1. \smallskip

For instance in Figure \ref{evoleng}, elitism is set to {\tt
  strong} and the number of elite parents to $10$: the best 10 parents
will anyway survive to next generation\footnote{Such a strategy is
generally used,  together with a weak selection
pressure, for instance when the fitness is very noisy, or is varying
along time.}.
Only the 90 worse individuals undergo the reduction toward
the {\tt Intermed.$\:$Pop.}$\:$---the 40 best out of
these 90 worse will survive this step--- and only 90 seats are
available in the {\tt New Pop.}, the 10 first seats being already
filled by the elite parents.

\subsection{Specifying the main evolutionary paradigms}
\label{paradigms}
The example in Figure \ref{evoleng} is typically a {\tt custom}
evolution engine. This section will give the parameter settings
corresponding to the most popular existing evolution engines
---namely
GAs (both generational and Steady-State), ES and EP.
Note that though those names correspond to the historical ``dialects,''
they are
used here to designate some particular combination of parameters,
regardless of any other algorithmic component (such as genotype and variation
operators).
However, going back to those familiar
engines by manually tuning the different parameters is  rather tedious.
This is the reason for the {\tt Evolution Engine} pull-out menu
on top of the panel: the user can specify in one click one of these
five ``standard'' engines, and
instantly see how this affects all the parameters.

After choosing one of the pre-defined engines, the user
can still modify all parameters. However, such modification will be
monitored, and as soon as the resulting engine
departs from the chosen one, the pull-out menu will
automatically turn
back to {\tt Custom}. The predefined engines are:\\

\noindent
{\bf Generational GA}
In that very popular algorithm, let $P$ be the population size.
$P$ parents are selected and give birth to $P$ offspring that in turn
replace the $P$ parents:
any selector is allowed, {\tt Fertile} size is equal to
{\tt Pop.} size, {\tt  Offspring}
size is set to {\tt Pop.} size, reduced {\tt N.E.P.} size to 0
(no parent should survive) and  {\tt Intermed. Pop.} size to {\tt
Pop.} size. In fact, none of the reducers is actually active in this
setting (this is a generational replacement).

Furthermore, weak elitism can be set (generally with size 1). {\tt
  Fertile} size can be reduced (this is equivalent to {\em truncation
  selection}) without leaving the {\tt GGA} mode.\\

\noindent
{\bf Steady-State GA}
In Steady-State GA, a single offspring is created at each generation, and
replaces one of the parent, usually chosen by tournament. Again, any
selector is allowed, {\tt Fertile} size is equal to
{\tt Pop.} size, but {\tt Offspring} size is set to 1 and the parents
are now reduced by some tournament reducer, to {\tt Pop.} size minus
one, while the final reducer is not active\footnote{
An ``age'' tournament
should be made available soon, as many SSGA-based algorithms do use
age as the criterion for the choice of the parent to be replaced.}.

The  number of
offspring can be increased without leaving the {\tt SSGA} mode:
there is no clear limitation of this mode, though the number of
offspring should be kept small w.r.t. {\tt Pop.} size.
And of course, setting any type of elitism here is a misconception.\\

\noindent
{\bf Evolution Strategies}
There are two popular evolution engines used in the traditional
Evolution Strategies algorithms,  the $(\mu,\lambda)-ES$ and the
$(\mu+\lambda)-ES$: in both engines, there is no selection step, and
all $\mu$ parents give birth
to $\lambda$ parents. The $\mu$ individuals of the new population are
deterministically chosen from the $\lambda$ offspring in the
$(\mu,\lambda)-ES$ and from the $\mu$ parents {\em plus} the
$\lambda$ offspring in the $(\mu+\lambda)-ES$.
Both these evolution engines set the selection to
{\tt Sequential}, the {\tt Offspring} size and the {\tt Reduce
  offspring} size to $\lambda$ (no reduction takes place
there) and the final reducer to {\tt sequential}. Therefore, choosing
$(\mu,\lambda)-ES$ sets the {\tt Reduce N.E.P.} size to $0$ while
choosing the $(\mu+\lambda)-ES$ sets it to {\tt Pop.} size.

The $(\mu+\lambda)-ES$ engine already is strongly elitist.
The $(\mu,\lambda)-ES$ engine is not, and elitism can be added ---but
it then diverges from the original ES scheme.\\

\noindent
{\bf Evolutionary Programming}
In traditional EP algorithms, $P$ parents generate $P$ offspring (by
mutation only, but this is not the point here). Then the $P$ parents
plus the $P$ offspring compete for survival. Setting {\tt EP} mode in the
Evolution Engine menu sets the {\tt Offspring} size to {\tt Pop.} size,
both reduced sizes  for {\tt N.E.P.} and {\tt Offspring} to {\tt Pop.}
size as well (no reduction here) and the final reducer to
{\tt EP tournament}.
Note that early EP algorithms sometimes used a deterministic
choice for the survivors ---this can be achieved by choosing the {\tt
  Sequential} final reducer.

Here again, elitism can be added, but this switches back to {\tt
  Custom} engine.

\section{Discussion and Perspectives}
\label{conclusion}

As already argued in section \ref{discussion}, the very first
limitation of the GUIDE evolution engine comes from the chosen EA
skeleton, that does not allow weird evolution engines. However, we
firmly believe that most existing EA application use some evolution
engine that falls in this framework.

The forthcoming improvements of this part of GUIDE are concerned with
adding new selector/reducer options, more specifically selection
procedures based on other criteria. The {\tt Age tournament} has
already been mentioned in the SSGA context. But all multi-objective
selection procedures will also be added (with additional options in
the {\tt Maximise/Minmise} pull-out menu.


Going away from the evolution engine, the asymmetry in terms of
flexibility between the Evolution Engine and the Algorithm
Visualisation Panels cries out for a graphical interface allowing the
user to specify the genome structure and the variation operators. Such
interface is not as utopian as it might seem at first sight: the
genome structure could be specified from basic types, and basic
constructors (e.g. heterogeneous aggregations, homogenous vectors,
linked lists, trees, \ldots). And there exist some generic ways to
design variation operators for such structures \cite{Surry:PPSN96}.

Of course, last but not least, after the Distribution Panel has been
adapted to ParaDisEO, the Experiment Monitor Panel should be
redesigned such that the user can graphically plot the evolution of any
variable in that window.

As a conclusion, the Evolution Engine Panel of GUIDE is only a first
step towards a widely available dissemination tool for EAs. But it already
achieves the demonstration
that all evolutionary algorithms are born equal if the user is provided with
enough parameters to tune.
GUIDE Evolution Engine Panel is a visual and pedagogical tool  that allows
one to understand
the intricacies of the different evolutionary paradigms.
While, quite often, Graphical User Interfaces reduce flexibility, GUIDE
offers at the Evolution Engine level readability, simplicity of use
and an easy way to experiment with complex parameters.


\bibliographystyle{splncs}

\end{document}